\RequirePackage{fix-cm}
\documentclass[twocolumn]{svjour3}          
\smartqed  
\usepackage{graphicx}
\usepackage{amsbsy,amssymb,latexsym,amsmath}
\usepackage{epstopdf}
\graphicspath{{.}}
\begin{document}

\title{Full-Body Collision Detection and Reaction with Omnidirectional Mobile Platforms: A Step Towards Safe Human-Robot Interaction}

\author{Kwan Suk Kim \and Travis Llado \and Luis Sentis}

\institute{K. Kim \and T. Llado \and L. Sentis \at The University of Texas at Austin \\ \email{\{kskim@, travisllado@, lsentis@austin\}.utexas.edu}
}

\date{Received: date / Accepted: date}

\maketitle

\begin{abstract}
In this paper, we develop estimation and control methods for quickly reacting to collisions between omnidirectional mobile platforms and their environment. To enable the full-body detection of external forces, we use torque sensors located in the robot's drivetrain. Using model based techniques we estimate, with good precision, the location, direction, and magnitude of collision forces, and we develop an admittance controller that achieves a low effective mass in reaction to them. For experimental testing, we use a facility containing a calibrated collision dummy and our holonomic mobile platform. We subsequently explore collisions with the dummy colliding against a stationary base and the base colliding against a stationary dummy. Overall, we accomplish fast reaction times and a reduction of impact forces. A proof of concept experiment presents various parts of the mobile platform, including the wheels, colliding safely with humans. 
\end{abstract}

\section{Introduction}
As mobile robots progress into service applications, their environments become less controlled and less organized compared to traditional industrial use. In these environments, collisions will be inevitable, requiring a thorough study of the implications of this type of interaction as well as potential solutions for safe operation. With this in mind, we are interested in characterizing the safety and collision capabilities of statically stable mobile bases moving in cluttered environments. The work presented here is the first of which we are aware to address, in depth, the mitigation of the effects of collisions between these types of sizable robots and objects or people.

The majority of work addressing mobility in cluttered environments has centered around the idea of avoiding collisions altogether. However, collisions between robotic manipulators and objects and humans have been investigated before \cite{Haddadin2009,Yamada1996}. Push recovery in humanoid robots allows them to regain balance by stepping in the direction of the push \cite{Pratt2006} or quickly crouching down \cite{Stephens-Thesis:2011}. Inherently unstable robots like ball-bots \cite{Nagarajan2013,Kumagai2008} and Segways \cite{Nguyen2004} have been able to easily recover from pushes and collisions using inertial sensor data. A four-wheel robotic base with azimuth joint torque sensors \cite{Fremy2010} has been able to respond to human push interactions, but only when its wheels are properly aligned with respect to direction of the collision. Also, a non-holonomic base with springs on the caster wheels was recently developed \cite{Kwon2011} and reported to detect pushes from a human, but with very preliminary results and without the ability to detect forces in all directions or detect contacts on the wheels themselves. In this work, we focus on non-stationary robots, as opposed to fixed base manipulators.
In the field of non-stationary robotic systems, such as statically or dynamically balancing mobile bases and legged robots, one of the key deficiencies is the availability of collision reaction methods that can be used across different platforms. Dynamically balancing mobile bases and humanoid robots rely on IMU sensing to detect the direction of a fall and then regain balance along that direction. However, this type of method is limited to robots which naturally tip over at the slightest disturbance.

The main objective of this paper is to develop general sensing and control methods for quickly reacting to collisions in statically stable mobile bases. 
Specifically, we develop methods that rely on joint level torque sensing instead of inertial measurement sensing to determine the direction and magnitude of the collision forces. If IMUs were used, accelerations would only be sensed accurately once the robot overcomes static friction which, for a sizable robot, could be quite large. Torque sensors, which are mounted next to the wheels, can quickly detect external forces sooner than IMUs and therefore are more suitable for quick collision response. Equally important is the fact that statically stable mobile bases can move in any direction or not at all in response to a collision, whereas dynamically balancing mobile bases and humanoid robots must move in the direction of the collision. This ability makes statically stable mobile bases more flexible when maneuvering in highly constrained environments.

To provide these capabilities, we take the following steps: (1) we develop a floating base model with contact and rolling constraints for an omnidirectional mobile base; (2) we process torque sensor signals using those models and statistical techniques; (3) we estimate roller friction and incorporate it into the constrained dynamics; (4) we implement a controller to quickly escape from the collisions; (5) we present an experimental testbed; and (6) we perform experiments including several calibrated collisions with the testing apparatus, and a proof of concept experiment in which the robot moves through a cluttered environment containing people against whom it must safely collide.

Overall, our contributions are (1) developing the first full-body contact sensing scheme for omnidirectional mobile platforms that includes all of the robot's body and its wheels, (2) being the first to use floating base dynamics with contact constraints to estimate contact forces, and (3) being the first to conduct an extensive experimental study on collisions with human-scale mobile bases.

\begin{figure}\centering
\includegraphics[width=0.8\linewidth, clip=true ] {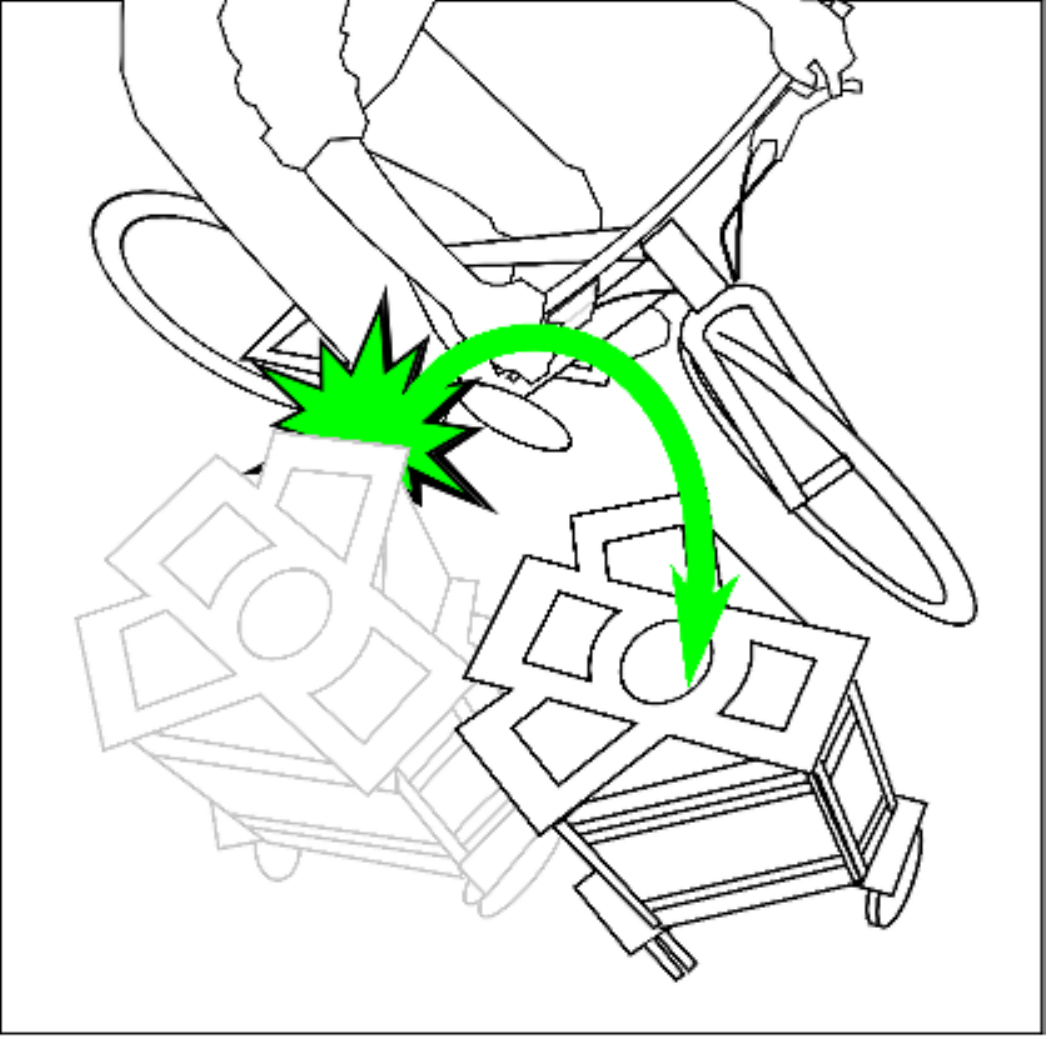}
\caption{{\bf Concept:} unexpected collision between a robot and a person on a bicycle, as presented in our supporting video.}
\label{concept}
\end{figure}

\section{Related Work}
\vspace{0.15in}

\subsection{Mobile Platforms with Contact Detection}

To be compliant to external forces, mobile robots have adopted various sensing techniques. One simple way to detect external forces is by comparing actual and desired positions \cite{Kim2013} or velocities \cite{Doisy2012}. This method is easy to implement because it can use the built-in encoders on the robot joints or wheels to detect external forces. However, the ability to detect contacts using this method depends largely on the closed-loop impedance chosen for the control law. 

Another means of detecting external forces is physical force\slash torque sensors such as strain gauges or optoelectronics. This approach has been used in many mobile platform applications such as anticipating user intention with a force-based joystick \cite{Sabatini2002}, developing a handle with force\slash torque sensing capabilities \cite{Spenko2006}, implementing an impedance control law based on force\slash torque sensed on a handle \cite{Chuy2007}, and quantifying user intent and responding with an admittance controller based on a force\slash torque sensor mounted on a stick \cite{Huang2008,Wakita2013}. However, all of these methods rely on detecting forces and torques at a specific location, such as on a handle, or joystick. When the user interacts or collides with other parts of the robot's body, such robots will not be able to respond to the applied forces safety. 

In \cite{Hirata2003}, a force\slash torque sensor measures forces between the mobile robot's body and an external protection cover, providing partial safety, but collisions against the wheels cannot be detected. In \cite{Fremy2010} they introduce a quasi omni-directional mobile robot that is compliant to external forces by measuring torques on the yaw joints of its caster wheels. This technique can detect collisions on the wheels like ours, but suffers from singularities which limit both the directions in which it can detect force and its freedom of motion. In \cite{Kwon2011} a sensorized spring system is installed on the frame of a mobile base with caster wheels and is used for push interactions. However, the base can respond to forces only in limited directions and is once more insensitive to collisions against the wheels. 

Other sensing properties have been used for contact interactions, notably the tilt measured by an inertial measurement unit on ball-bot robots \cite{Nagarajan2013}. This type of robot, and the associated inertial sensing, have been used effectively to handle contact interactions with people \cite{Kumagai2008}. However, the main drawback of this method is that the robot must move in the direction of the disturbance or it will fall over. In contrast, non-inertial force sensing techniques like ours allow a robot to react in any direction upon collision or force interaction. This ability might be very useful when producing planned movements tailored to the external environment. 

\subsection{Contact Detection via Joint Torque Sensing}

Several existing studies use joint torque sensing to detect contact, like us, but only address serial robotic manipulators. Note that this technique is distinct from the commonly used multi-axis force\slash torque sensor located at the end effector of a manipulator. Many researches have investigated sensing external forces on all parts of a manipulator's body using distributed joint torque sensors~\cite{Wu1985,Luca2006}. 

Like our method, this indirect external force sensing requires estimation that considers dynamic effects such as linkage and motor masses, inertias, momentum, gravitational effects, and friction. Statistical estimation methods~\cite{Fang2011} are used to estimate external forces based on joint torque sensing~\cite{Le2013}. These methods have inspired our research, but we note that we have taken similar approaches for a mobile platform instead of for a robotic manipulator. A mobile platform has different dynamics because it has a non-stationary base and its wheels are in contact with the terrain. Such differences imply different dynamic models and modifications of the estimation methods.

\subsection{Safety Analysis in Robotics}

Pioneering work on safety criteria for physical human-robot interaction are provided in~\cite{Yamada1996}. In particular, curves of maximum tolerable static forces and dynamic impacts on various points of the human body are empirically derived. A method to detect external forces using motor current measurements and joint states is proposed, and a viscoelastic skin is utilized to dampen impacts.

In \cite{Zinn2004} the positive effects on safety of actuators with a series elastic compliance are brought up but linked to lower performance. A double macro-mini actuation approach is proposed to accomplish safe operation while maintaining performance, and the automotive industry's Head Injury Criterion index is used to demonstrate the benefits of this approach in terms of safety.

A comprehensive experimental study on human-robot impact is conducted in \cite{Haddadin2009}. This study suggests that the Head Injury Criterion is not well suited for studying injuries resulting from human-robot interaction. Instead, the authors propose contact forces acting as a proxy to bone fractures as their injury indicator. The low output inertia achievable with their torque control manipulators is shown to be highly conducive to preventing injury during collisions.

Also relevant to our work is the study considering child injury risks conducted in \cite{Fujikawa2013}. Extensive experimental data is obtained from a 200Kg mobile robot moving at speeds of 2Km/h and 6Km/h and colliding against a robot child dummy fixed to a wall. The head injury and neck injury criteria are used to study the consequences of the impacts, and the severity of injury is expressed by the Abbreviated Injury Scale. Those criteria are reinforced with analysis of chest deflection for severity evaluation. In contrast with our work, their mobile platform is uncontrolled and does not have the ability to sense contact. This study is focused purely on impact analysis instead of contact sensing and safe control.

\subsection{Model-Based Control of Omnidirectional Platforms}

A mobile platform colliding or interacting with the environment is not only affected by external forces, but also by static and dynamic effects such as the robot's inertia, its drivetrain and wheel friction, and other mechanical effects. \cite{Zhao2009} considers a simulated system consisting of a 6-DOF omni-directional mobile robot with caster wheels, and addresses the modeling and control of motion and internal forces in the wheels. \cite{Djebrani2012} derives the dynamic equation including the rolling kinematic constraint for a mobile platform similar to ours, but uses an oversimplified dynamic friction model with respect to the effects of roller friction. Studies that incorporate static friction models include~\cite{Viet2012,BarretoS.2014}, but again these use oversimplified models that ignore omniwheel and roller dynamics. The studies above are mostly theoretical, with few experimental results.

\section{System Characterization}
\label{sec:syschar}

\subsection{Hardware Setup}

To perform experimental studies on human-robot collisions, we have built a series of capable mobile platforms. This study uses the most recent.
We began designing mobile bases to provide omnidirectional rough terrain mobility to humanoid robot upper bodies~\cite{sentis2012rss}. The newest iteration of our platform, produced in~\cite{Kim2013}, replaced the previous drivetrain with a compact design that minimized backlash by using belts and pulleys. Rotary torque sensors in the drivetrain and harmonic drives on the actuators were incorporated into the base in ~\cite{Kim2014}, enabling accurate force feedback control for impedance behaviors. The electronics in the current system improve over that of ~\cite{Kim2014} in that the once centralized torque sensor signal processing is now divided into each actuator's DSP in order to minimize electrical crosstalk. This paper is the first study that uses the torque sensors on the hardware base for full-body model-based estimation of the contact forces.

Rotary torque sensors in the wheel drivetrains produce the unique feature of our base: full-body contact estimation on all parts of its body, including any part of the wheels. An alternative would have been to cover all of a robot's body with a sensitive skin, but this option would have left the wheels uncovered and therefore unable to detect contact. We note that the wheels are often the first part of the base that collides with unexpected objects. Therefore, our solution with three rotary torque sensors in the wheel's drivetrain is the first and only one of which we are aware that can respond to collisions on all parts of the mobile platform. Additionally, the harmonic drives and belt-based drivetrain of the base minimize backlash and therefore achieve more accurate force sensing. 

\begin{figure}\centering
\includegraphics[width=\linewidth, clip=true ] {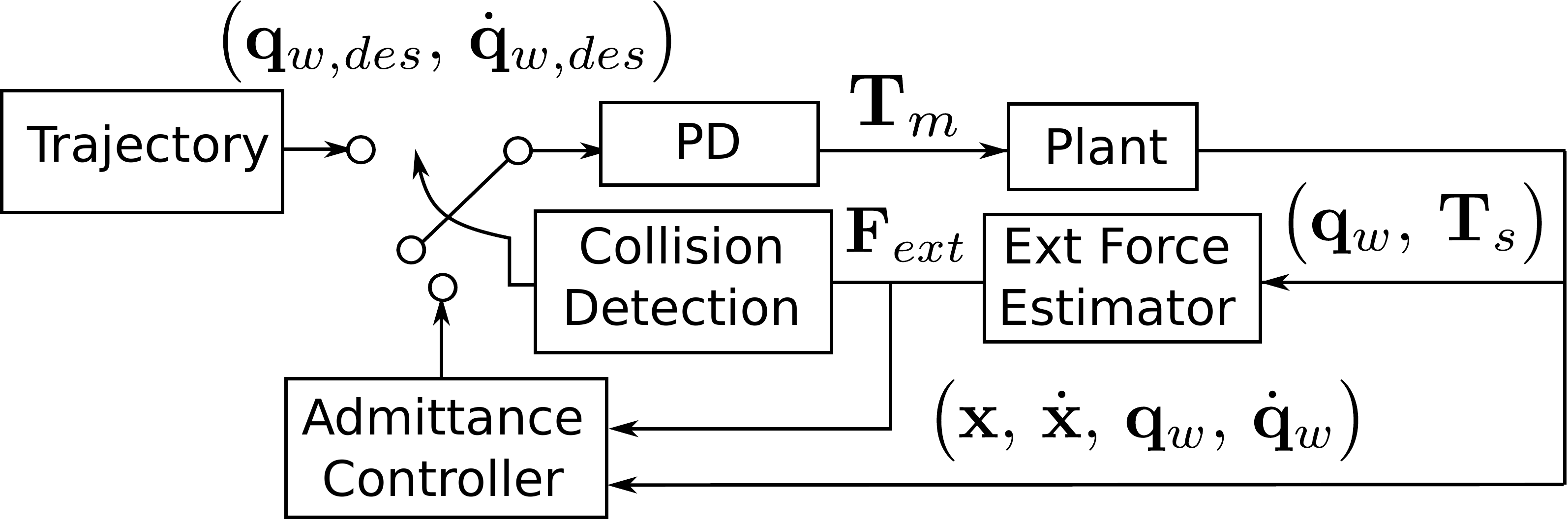}
\caption{{\bf Control Diagram} showing how estimated external force, $\mathbf{F}_{ext}$, is fed through the collision detector and ultimately determines the position controller's input. When the robot detects a collision it uses an admittance controller in place of its usual trajectory to escape the contact as fast as it can safely move.}
\label{controller}
\end{figure}

\subsection{Safety Controller Design}

When a mobile base collides with people, two cases can be previously distinguished: In unconstrained collisions a person can be pushed away, whereas in fixed collisions the person is pushed against a wall. In either scenario our robot moves away from the collision as quickly as possible to mitigate injury. 

Fig. \ref{controller} shows our proposed control architecture for detection of and reaction to collisions. Under normal circumstances, the controller tracks a trajectory given by a motion planner or sensor-based algorithm. When an external force breaches our contact threshold, the controller switches on an admittance controller. This admittance controller generates a trajectory that responds to the sensed external force and rapidly leads the robot away from the contact. We tested both an impedance and an admittance controller in this role during the course of our research, but found the admittance controller to be more responsive.

\begin{figure*}
\begin{center}
\includegraphics[width=0.8\linewidth, clip=true]{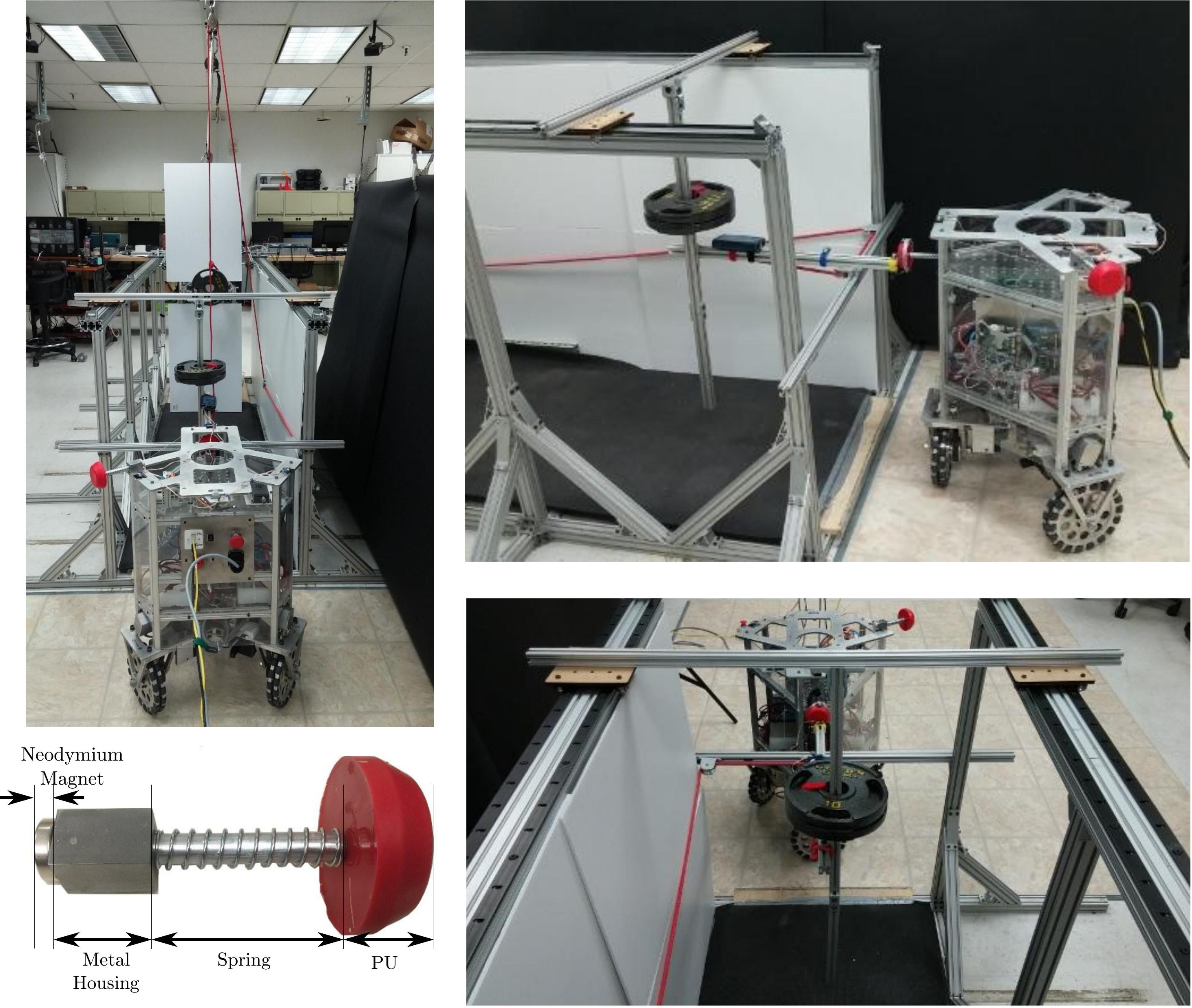}
\end{center}
\caption{{\bf Collision Testing Apparatus} simulates human contact using a 10kg mass on a slider. This one degree of freedom system is accelerated via a second weight hanging from an elaborate pulley system, and can also be used to apply a static force. Motion capture markers attached to the slider and the PU bumper are used to measure their position.}
\label{setup}
\end{figure*}

\subsection{Reaction to Collisions}

The admittance controller is designed to provide compliance with respect to the external force. The desired dynamics can be expressed as
\begin{align}\label{eq:simpledyn}
M_{des} \ddot{x} + B_{des} \dot{x} = F_{ext,x}(t),
\end{align}
where $M_{des}$ and $B_{des}$ are the desired mass and damping of a virtual compliant system, and $F_{ext,x}(t)$ is the time dependent force disturbance applied to the system. Assuming the external force is close to a perfect impulse, i.e. a Dirac delta function, the above equation can be solved to produce the desired trajectory,
\begin{equation}\label{eq:traj}
x\left(t\right) = x_0 + \frac{F_{ext,x}}{B_{des}} \left( 1 - e^{-B_{des}/M_{des} \, t }\right),
\end{equation}
where $x_0$ is the position of the system when the collision happens.
An identical admittance controller operates on the $y$ degree of freedom.

Our controller attempts to maintain constant yaw throughout the collision, i.e.
\begin{equation}
\theta(t) = \theta_0.
\end{equation}
Combining the three degrees of freedom, we write the robot's full trajectory as
\begin{equation}
\mathbf x_{des}(t) = \big(x_{des}(t), \;\;\; y_{des}(t), \;\;\; \theta_{des}(t) \big)^T.
\end{equation}
This trajectory is differentiated and then converted into a desired joint space trajectory using the constrained Jacobian, $\mathbf J_{c,w}$ given in Eq. (\ref{eq:jcw}), i.e.
\begin{gather}
\dot{\mathbf q}_{w,des}(t) = \mathbf J_{c,w}(t) \; \dot {\mathbf x}_{des}(t),\\[1mm]
{\mathbf q}_{w,des}(t) = {\mathbf q}_{w,des}(t_0) + \int_{t_0}^t \dot{\mathbf q}_{w,des}(\tau) d\tau,
\end{gather}
and fed to the PD controller of Fig. \ref{controller} to achieve the intended impedance behavior.

\subsection{Collision Testbed}

To assess the safety of our mobile platform, we constructed a calibrated collision testbed. Following the collision test procedure used in the automotive industry \cite{UNECE2011}, we chose a 10kg mass as our leg-form test dummy. The collision dummy is attached to a sliding system which provides a single degree of freedom for impact, and is accelerated by a free falling weight. In Fig. \ref{setup} we illustrate details of the test environment. The absolute positions of the dummy and the mobile base are measured by the Phase Space motion capture system described in \cite{Kim2013}. Four markers on the mobile base measure its position and two markers on the dummy measure its linear motion. 

\subsection{Stiction-Based Bumper}
\label{material}

The\, time \, requirement\,  for our base to detect collision and reverse direction is roughly one hundred milliseconds. Keeping the collision time brief works to reduce injury, but is insufficient to eliminate it\, altogether.\  Though it is impractical to fully pad a robot, some padding can drastically reduce the collision forces due to collision with specific parts of the robot's body. Yet reducing the forces makes the problem of detecting the collision more difficult, and increases the amount of time before the robot acknowledges an impact. We have designed a one DOF springloaded bumper with a relatively long travel to study the design of safe padding for omnidirectional robots. This design features a magnetic lock at peak bumper extension, which works to allow earlier detection of a collision, while simultaneously reducing the overall maximum impact force. Details of the bumper can be found in Fig. \ref{setup}.


\begin{figure}\centering
\includegraphics[width=\linewidth, clip=true ] {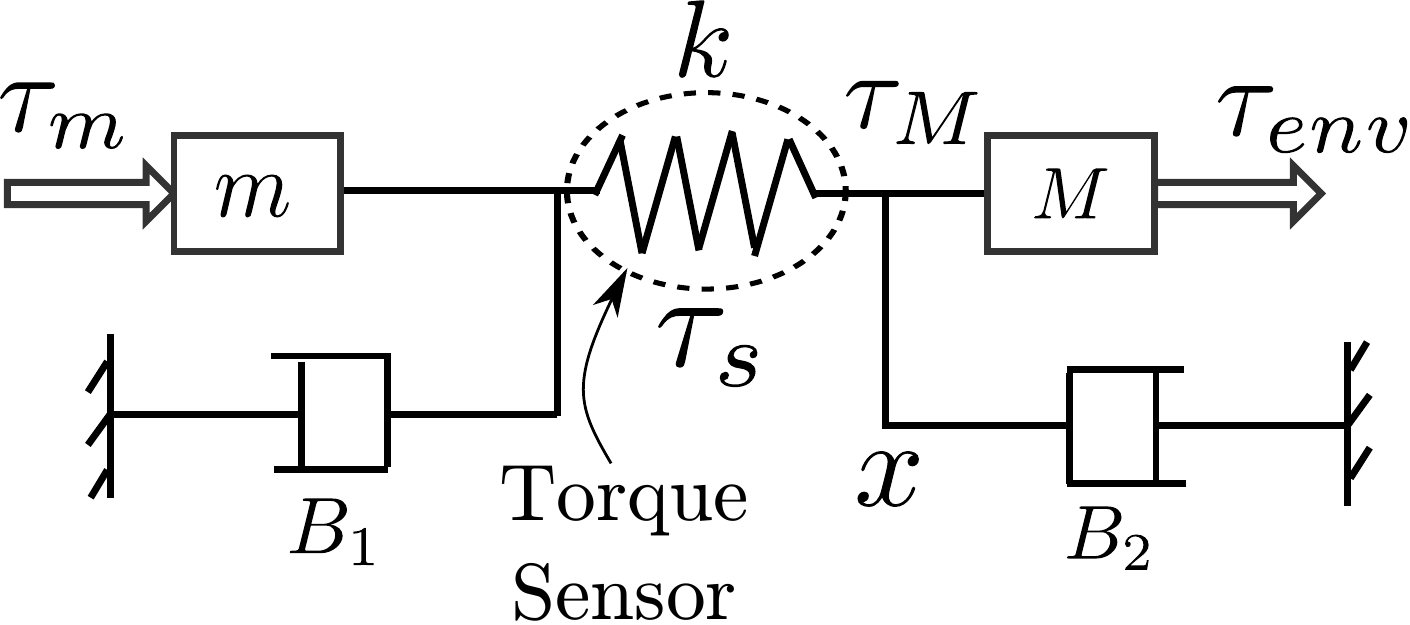}
\caption{{\bf Actuator Model} including the torque sensor, modeled as a spring. The two masses $m$ and $M$ represent the motor inertia, reflected through the gear system, and the load mass. Motor side friction and load side friction are expressed as the damping terms $B_1$ and $B_2$, respectively.}
\label{model} 
\end{figure}

\section{Full-Body External Force Estimation}
\label{sec:forceestimation}
To estimate external forces based on drivetrain torque sensing, we rely on a model of the actuators, and on the robot's kinematics and dynamics. The constrained kinematic mapping between the base's motion and wheel motion is used to find the base and omniwheel roller velocities based on measured wheel velocities. The actuator model provides a mapping between motion and expected torque sensor values in the absence of external forces. This model is trained empirically to better estimate the friction in the omni-wheel rollers. Ultimately the position, magnitude, and direction of the applied external forces is estimated based on the deviation of the observed wheel torques from those predicted by the force free model, and the kinematics are again invoked to transform this into the Cartesian frame.

To build an intuition of our method for estimating external forces, consider the single actuator system shown in Fig. \ref{model}. In this system, the torque sensor is modeled as a torsional spring, with spring constant $k$, and its displacement is proportional to the torque applied to the sensor. The spring is compressed or extended through the combined action of the motor, the wheel's inertia, and the external environment. Some of the important variables include the motor's torque, $\tau_m$, its rotor's mass, as reflected through mechanical gearing, $m$, the gear friction, $B_1$, the load's mass (i.e. the wheel, or the robot itself in the constrained case), $M$, the friction between the wheel and the external environment, $B_2$. But most importantly, the torque $\tau_{env}$ includes the effect of the wheel traction on the floor and any possible external collision with objects or people,
\begin{equation}
\tau_{env} = \tau_{trac} + \tau_{ext}.
\end{equation}
Assuming that the effect of the wheel traction, $\tau_{trac}$ can be modeled, our goal is to estimate the external forces, $\tau_{ext}$, based on observed sensor torque $\tau_s$:
\begin{equation}\label{extest}
\tau_{ext} = -\tau_{trac} + B_2 + M \ddot x - \tau_s.
\end{equation}
This method can then be applied to the estimation problem of the full base by using the kinematic constraint relationships between the wheels and the ground.

\subsection{Torque Output Dynamics}

To derive wheel and roller kinematics, we consider a planar scenario where the wheel moves omnidirectionally on a flat floor. In~\cite{Kim2013} we developed the following equations relating the contribution of the $i^{th}$ wheel's angular velocity, $\dot q_{w,i}$, and their omniwheel roller's angular velocity, $\dot q_{r,i}$, to the Cartesian velocity of the robot with respect to a fixed inertial frame, $\dot x$ and $\dot y$:
\begin{gather}
\dot{x} = r_r \dot{q}_{r,i} \cos \left( \theta + \phi_i \right) - \left( r_w \dot{q}_{w,i} - R \dot{\theta} \right) \sin \left( \theta + \phi_i \right),\label{const_eq1}\\[-2mm]
\dot{y} = r_r \dot{q}_{r,i} \sin \left( \theta + \phi_i \right) + \left( r_w \dot{q}_{w,i}  - R \dot{\theta} \right) \cos \left( \theta + \phi_i \right).\label{const_eq2}
\end{gather}
Where, $\theta$ is the absolute orientation of the robot's body, $R$ is the distance from the center of the robot's body to the center of the wheel, $r_w$ and $r_r$ are the radii of the wheels and their passive rollers, respectively, and $\phi_i$ is the angle from a reference wheel to the i-th wheel in sequential order, i.e. 0$^\circ$, 120$^\circ$, or 240$^\circ$. The kinematics of $\dot q_{w,i}$ and $\dot q_{r,i}$ are obtained from Eq. (\ref{const_eq1}) 
\begin{align}
&r_w \dot{q}_{w,i}= - \dot{x} \sin\left( \theta + \phi_i \right) + \dot{y} \cos \left( \theta + \phi_i \right) + R \dot{\theta}, \label{const_eq3}\\[1.5mm]
&r_r \dot{q}_{r,i}=\dot{x} \cos \left( \theta + \phi_i \right) + \dot{y} \sin \left( \theta + \phi_i \right). \label{const_eq4}
\end{align} 
Expressing these equations in matrix form,
\begin{gather}\label{eq:dotqw}
\mathbf{\dot q}_w = \mathbf J_{c,w} \, \mathbf{\dot x},\\[1.5mm]\label{eq:dotqr}
\mathbf{\dot q}_r = \mathbf J_{r,w} \, \mathbf{\dot x}
\end{gather}
where
\begin{gather}\label{eq:jcw}
\mathbf{J}_{c,w} \triangleq 
\frac{1}{r_w}
\begin{pmatrix}
-\sin\left( \theta + \phi_0 \right) &&& \cos\left( \theta + \phi_0 \right) &&& R \\
-\sin\left( \theta + \phi_1 \right) &&& \cos\left( \theta + \phi_1 \right) &&& R \\
-\sin\left( \theta + \phi_2 \right) &&& \cos\left( \theta + \phi_2 \right) &&& R
\end{pmatrix} \in \mathbb{R}^{3\times 3}, \\\label{eq:jcr}
\mathbf{J}_{c,r} \triangleq 
\frac{1}{r_r}
\begin{pmatrix}
\cos\left( \theta + \phi_0 \right) &&& \sin\left( \theta + \phi_0 \right) &&& 0 \\
\cos\left( \theta + \phi_1 \right) &&& \sin\left( \theta + \phi_1 \right) &&& 0 \\
\cos\left( \theta + \phi_2 \right) &&& \sin\left( \theta + \phi_2 \right) &&& 0
\end{pmatrix}\in\mathbb{R}^{3\times 3},
\end{gather}
are the Jacobian matrices, $\mathbf{q}_w \triangleq (q_{w,0}, \;q_{w,1}, \; q_{w,2})^T$, $\mathbf{q}_r \triangleq (q_{r,0}, \;q_{r,1}, \; q_{r,2})^T$, and $\mathbf{x} \triangleq (x, \; y, \; \theta)^T$. The system's generalized coordinates combine the wheel and Cartesian states
\begin{equation}
\mathbf{q} \triangleq
\begin{pmatrix}
\mathbf{x}^T & \mathbf{q}_{w}^T & \mathbf{q}_{r}^T 
\end{pmatrix}^{T}.
\end{equation}
Notice that we not only include wheel rotations, $\mathbf q_w$, but also side roller rotations, $\mathbf q_r$. This representation contrasts previous work on modeling that we did in~\cite{Sentis2013}. The main advantage, is that the augmented model will allow us to take into account roller friction which is significant with respect to actuator friction. As such, we will be able to estimate external interaction forces more precisely.

The mappings given in Eqs. (\ref{eq:dotqw}) and (\ref{eq:dotqr}) can be written as the constraint
\begin{equation}
\mathbf{J}_c \ \dot{\mathbf{q}} = 0,
\end{equation}
with 
\begin{equation}
\mathbf{J}_c \triangleq 
\begin{pmatrix}\label{eq:jcw}
\mathbf{J}_{c,w} &&& 
-\mathbf{I} &&&
\mathbf{0} \\
\mathbf{J}_{c,r} &&&
\mathbf{0} &&& 
-\mathbf{I}
\end{pmatrix}\in \mathbb{R}^{6\times 9}.
\end{equation}
Using the above kinematic constraints, one can express the coupled system dynamics in the familiar form
\begin{equation}\label{sys_dyn}
\mathbf{A} \ddot{\mathbf{q}} + \mathbf{B} + \mathbf{J}_c^T \boldsymbol\lambda_c = \mathbf{U}^T \mathbf{T}, 
\end{equation}
where $\mathbf A$ is the mass/inertia generalized tensor, $\mathbf B$ is a vector containing the estimated wheel drivetrain friction and roller to floor friction, and $\boldsymbol \lambda_c$ is the vector of Lagrangian multipliers associated with the traction forces of the wheel, where $\boldsymbol\lambda_{c,w}$ enforces the relationship between Cartesian robot position and wheel angle, and $\boldsymbol\lambda_{c,r}$ enforces the relationship between Cartesian robot position and omniwheel roller angle. In other words 
\begin{equation}
\boldsymbol\lambda_c = \left(\boldsymbol\lambda_{c,w}^T, \; \boldsymbol\lambda_{c,r}^T\right)^T.
\end{equation}
Additionally, $\mathbf U$ is the vector mapping motor torques to generalized forces, and $\mathbf T\in \mathbb{R}^3$ is the vector of output torques on the wheels. As mentioned previously, these are equivalent to the sensed torques,
$
T_s = T.
$
Values for the aforementioned matrices are
\begin{gather}\label{eq:A}
\mathbf{A} =
\begin{pmatrix}
\mathbf{M} & \mathbf{0} & \mathbf{0} \\
\mathbf{0} & I_w \mathbf{I} & \mathbf{0} \\
\mathbf{0} & \mathbf{0} & I_r \mathbf{I}
\end{pmatrix} \in \mathbb{R}^{9\times9}, \;\;\;
\mathbf{M} = 
\begin{pmatrix}
M & 0 & 0 \\
0 & M & 0 \\
0 & 0 & I_b \\
\end{pmatrix},\\[4mm]\label{eq:B}
\mathbf{B} =
\begin{pmatrix}
\mathbf{0} & 
\mathbf{B}_w^T & 
\mathbf{B}_r^T 
\end{pmatrix}^T\in \mathbb{R}^9,\quad
\mathbf{U} =
\begin{pmatrix}
\mathbf{0} & \mathbf{I} & \mathbf{0}
\end{pmatrix}\in \mathbb{R}^{3\times 9},\;
\end{gather}
where $M$, $I_b$, $I_w$, and $I_r$ are the robot's mass, body inertia, wheel inertia, and roller inertia respectively. The damping term, $\mathbf{B}$, consists of the damping at the wheel output (i.e. torque sensor bearings and belt drive), $\mathbf{B}_w$, and the damping from the side rollers, $\mathbf{B}_r$. 
We note that the side rollers do not have bearings and consist of a relatively high friction bushing mechanism. Therefore, the wheel friction is negligible relative to that of the side rollers. Thus we estimate only roller friction in our final controller. Eq.~(\ref{sys_dyn}) can be decomposed into separate equations expressing robot's body, wheel and roller dynamics as
\begin{gather}\label{eq:system}
\begin{cases}
\mathbf{M} \,
\ddot{\mathbf{x}}
+
\begin{pmatrix}
\mathbf{J}_{c,w}^T & \mathbf{J}_{c,r}^T
\end{pmatrix} \boldsymbol\lambda_c 
= \mathbf{0},\\[2mm]
I_w
\ddot{\mathbf{q}}_w
- \boldsymbol\lambda_{c,w}
= \mathbf{T},\\[2mm]%
I_r
\ddot{\mathbf{q}}_r
+ \mathbf{B}_r -\boldsymbol\lambda_{c,r}
= \mathbf{0}.
\end{cases}
\end{gather}
Using the second and third equations above, we can calculate the constraint forces on the wheels and rollers,
\begin{align}
\boldsymbol\lambda_c &= 
\begin{pmatrix}
I_w
\ddot{\mathbf{q}}_w 
- \mathbf{T} \\[2mm]
I_r
\ddot{\mathbf{q}}_r
+ \mathbf{B}_r
\end{pmatrix}.\label{lambda_c}
\end{align}
Substituting this expression into the first equation of the equation system~(\ref{eq:system}) we get
\begin{multline}\label{sys_dyn04}
\mathbf{M}\,
\ddot{\mathbf{x}}
+
\mathbf{J}_{c,w}^T 
\left(
I_w
\ddot{\mathbf{q}_w} 
- \mathbf{T}
\right)
+ \mathbf{J}_{c,r}^T
\left(
I_r
\ddot{\mathbf{q}_r}
+ \mathbf{B}_r
\right)
= \mathbf{0}.
\end{multline}
Solving the above for the output torque, $T$, we get the nominal torque model
\begin{align}\label{f_ext03}
\mathbf{T} &=
\mathbf{J}_{c,w}^{-T}
\Big[
\mathbf{M}\,
\ddot{\mathbf{x}}
+ \mathbf{J}_{c,r}^T
\left(
I_r \mathbf{\ddot{q}}_r +
\mathbf{B}_r
\right)
\Big]
+ I_w \mathbf{\ddot{q}}_w.
\end{align}
This model predicts torque sensor values in the absence of external forces. By comparing the torque sensor data against this estimate, as in Eq. (\ref{extest}), we will be able to infer the external forces. But first we must calibrate the roller friction estimate.

\subsection{Empirical Estimation of Roller Damping}

As we shown in Eqs. (\ref{sys_dyn}) and (\ref{eq:B}), the\, damping\, terms\, associated\, with\, the\, output\, dynamics\, correspond\, to\, wheel output damping, $\mathbf B_w$ and roller damping, $\mathbf B_r$. Wheel output damping consists of the friction sources between the torque sensor and the wheel, which correspond to sensor bearings and the belt connecting the sensor to the wheel. Notice that gear friction is not included, as the torque sensor is located after the gears. When we lift the robot of the ground and rotate the wheels, the mean value of the torque sensor signal is close to zero, meaning that the drivetrain output friction is negligible compared to roller friction. On the other hand, roller friction is relatively large as the rollers do not have bearings and therefore endure high friction when rotating in their shaft. In the next lines we will explain our procedure to estimate roller damping based on torque sensor data.

\begin{figure*}[ht] \centering
\includegraphics[width=160mm, clip=true]{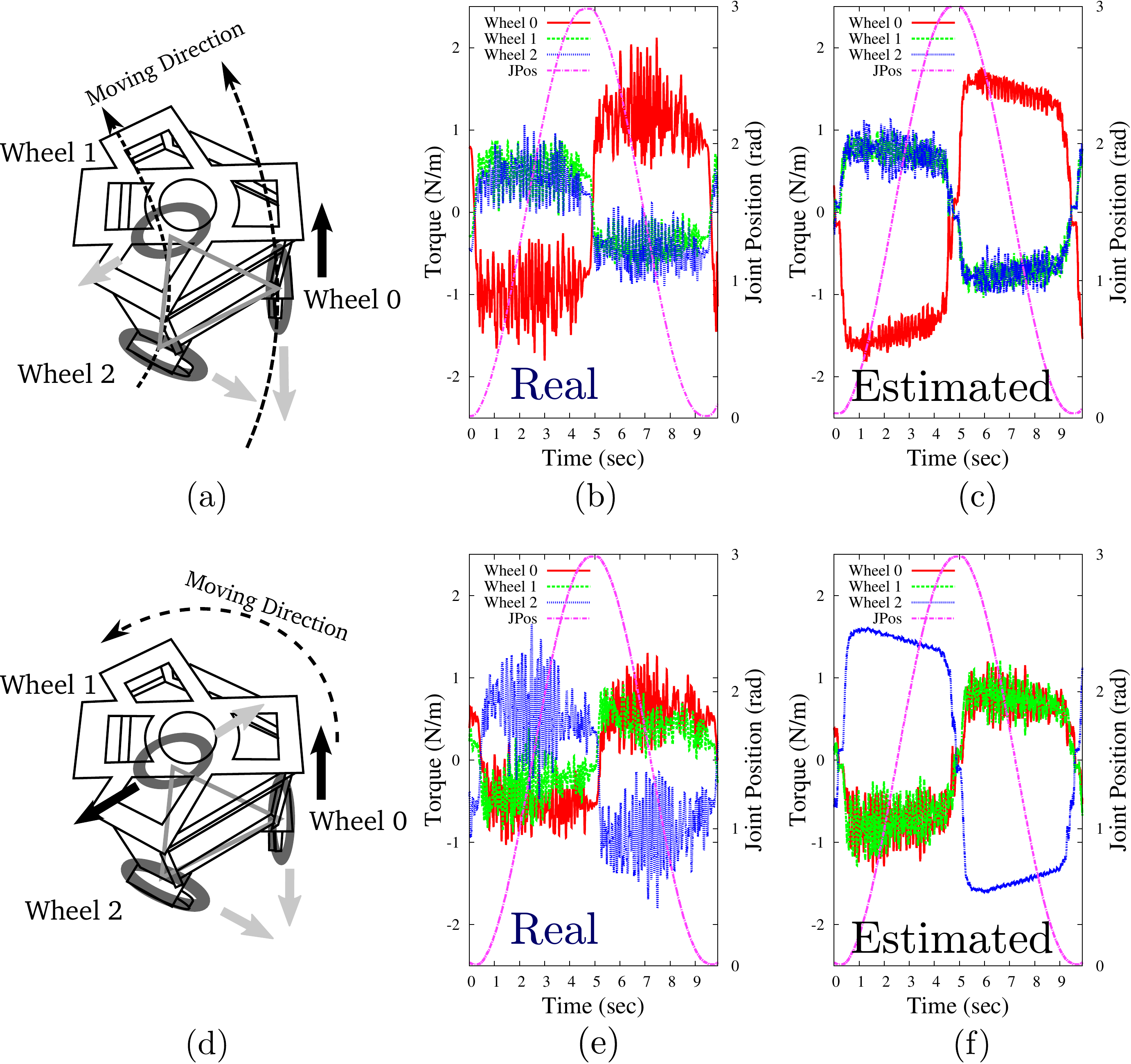}
\caption{{\bf Torque Signals from Simple Motions}\label{roller_friction} are used to calibrate the roller friction model. No external forces are applied to the robot in this test. The JPos lines represent the motion along the two simple arc trajectories. Torque signals from the calibrated model are shown to the right of the graphs representing the actual data on which they were trained. Subfigures (a-c) represent the rotation of the robot about a virtual pivot outside the base of support, while (d-f) show a pivot centered on Wheel~2. Gray arrows in figures (a) and (d) represent the torque sensed at the wheels, while the black arrows represent wheel motion. By comparing (b) against (c) and (e) against (f), we can conclude that the expected roller friction torque model at least partially captures the gross shape of the data.
}
\label{roller_friction}
\end{figure*}

Fig.~\ref{roller_friction} demonstrates the two experiments under which the roller friction model was calibrated. In these tests, joint position controllers for each wheel, simple high gain servos, push the robot through a nominal path, and the resulting torque sensor values are measured in the absence of any external force. In Subfig. \ref{roller_friction} (a) we show an experiment in which wheel 0 moves sinusoidally with time while the other two wheels remain fixed, resulting in an arc motion of the entire robot. In Subfig. \ref{roller_friction} (b) we plot the sinusoidal joint trajectory of wheel 0 and the torque sensor readings from the three wheels. The torque signals on all wheels show an approximately square wave shifting phase according to the direction of wheel's 0 motion. Because of this pattern, we assume that most of the friction is due to Coulomb effects instead of dynamic friction effects. We approximate this Coulomb friction in our model using a $tanh$ softening of the signum function, i.e.
\begin{align}\label{eq:br}
B_{r,i} = B_r \tanh \left( \alpha \, \dot{q}_{r,i}  \right),
\end{align}
where the magnitude $B_r$ and scaling factor $\alpha$ are tunable parameters that we adjust based on the empirical data. To do the tuning, we implemented Eq. (\ref{f_ext03}) in a software simulation and compared its output to the experimental data. In that equation, the accelerations of the wheels, the robot's body and the side rollers must be known. We calculate them using the wheel trajectories, 
$q_{w,0} = 3/2 - 3/2 {\rm cos}(2 \pi \omega t)$, $q_{w,1} = q_{w,2} = 0$, which can be easily differentiated twice to obtain $\ddot{\mathbf q}_w$. To obtain the robot's body acceleration we use the inverse of Eq. (\ref{eq:dotqw}) and take the second derivative, yielding 
\begin{equation}\label{eqddotx}
\ddot{\mathbf{x}} = \mathbf J_{c,w}^{-1} \ddot{\mathbf q}_{w} + \dot{\mathbf J}_{c,w}^{-1} \, \dot{\mathbf q}_{w}.
\end{equation}
Eq. \ref{eq:dotqr} then provides the roller accelerations 
\begin{equation}
\ddot{\mathbf q}_r = \mathbf J_{r,w} \ddot{\mathbf x} + \dot{\mathbf J}_{r,w} \dot{\mathbf x}.
\end{equation} 
Plugging these values into the simulation of Eq. (\ref{f_ext03}), with $\mathbf B_r$ given by the model of Eq. (\ref{eq:br}) we searched over $B_r$ and $\alpha$ until the simulation matched the real data. In Fig. \ref{roller_friction} (c) we show the result of the simulation using Eq. (\ref{f_ext03}) which can be compared to the real data of Fig. \ref{roller_friction} (b). Our final model parameters were $B_r = 0.2 Nm$ and $\alpha = 0.4$. 

To further validate the procedure we conducted a second estimation process, shown in the Figs. \ref{roller_friction} (d), (e) and (f) in which two wheels of the mobile base track a sinusoidal trajectory while one of them remains at a fixed joint position. As we can see, the simulated torques with the estimated roller friction model of Fig. \ref{roller_friction} (f) has a good correspondence to the actual data of Fig. \ref{roller_friction} (e).

\subsection{Model-Based Force Estimation}
Following the simplified estimation of external torques from Eq. (\ref{extest}), we modified Eq. (\ref{sys_dyn}) to account for external forces, yielding
\begin{align}\label{ext_dyn}
\mathbf{A} \ddot{\mathbf{q}} + \mathbf{B} + \mathbf{J}_c^T \boldsymbol\lambda_c + \mathbf{J}_{ext}^T \mathbf{F}_{ext} = \mathbf{U}^T \mathbf{T}. 
\end{align}
where $\mathbf{J}_{ext}$ is the Jacobian corresponding to the location of the external forces, and $F_{ext}$ is an external wrench containing a Cartesian force and a torque, i.e.
\begin{align}\label{f_ext}
\mathbf{F}_{ext} \triangleq
\begin{pmatrix}
F_{ext,x} & F_{ext,y} & \tau_{ext}
\end{pmatrix}^T.  
\end{align}
The differential kinematics of the point on the exterior of the body at which the external force is applied can be expressed in terms of the robot's differential coordinates as
\begin{equation}\label{eq:dotxext}
\dot{\mathbf{x}}_{ext} = 
\dot{\mathbf{x}} + \dot{\theta}\, \mathbf{i}_z \times {\mathbf{d}}
= \mathbf{J}_{ext,b} \dot{\mathbf{x}}
\end{equation}
where 
$\mathbf{x}_{ext} \triangleq \begin{pmatrix} x_{ext} & y_{ext} & \theta_{ext}\end{pmatrix}^T$,  $\dot \theta$ is the angular velocity of the base, $i_z$ is the unit vector in the vertical, $z$, direction, $\times$ is the cross product, and $\mathbf{d}$ is a vector describing the distance from the center of the robot to the collision point. Developing the above equations, we can define
\begin{equation}\label{eq:jext}
\mathbf J_{ext,b} \triangleq
\begin{pmatrix}
1 &&& 0 &&& y - y_{ext} \\
0 &&& 1 &&& x_{ext} - x \\
0 &&& 0 &&& 1 
\end{pmatrix}\in\mathbb{R}^{3\times 3}.
\end{equation}
Extending Eq. (\ref{eq:dotxext}) with respect to the full generalized coordinates yields
\begin{equation}
\dot{\mathbf{x}}_{ext} = 
\mathbf{J}_{ext} \, \dot{\mathbf{q}}, \quad {\rm with} \quad
\mathbf{J}_{ext} \triangleq
\begin{pmatrix}
\mathbf{J}_{ext,b}
&
\mathbf{0}_{3\times 6}
\end{pmatrix}.
\end{equation}
Using the above expression for $J_{ext}$ in the extended dynamics of Eq. (\ref{ext_dyn}), and neglecting the effect of the wheel and roller inertias, $I_w\approx 0$, and $I_r \approx 0$ with respect to the robot's mass, and the effect of the wheel friction, $\mathbf B_w \approx 0$ with respect to the roller friction, we get a similar system of equations than that shown in Eqs. (\ref{eq:system}), i.e.
\begin{gather}
\begin{cases}
\mathbf{M} \,
\ddot{\mathbf{x}} + \Big( \mathbf{J}_{c,w}^T\;\; \mathbf{J}_{c,r}^T \Big)
\boldsymbol\lambda_c + \mathbf{J}_{ext,b}^T F_{ext} = \mathbf{0},\\[2mm]
- \boldsymbol\lambda_{c,w} = \mathbf{T},\\[2mm]
\mathbf{B}_r -\boldsymbol\lambda_{c,r}
= \mathbf{0}.
\end{cases}
\end{gather}
Substituting $\boldsymbol{\lambda}_c \triangleq (\mathbf \lambda_{c,w}, \mathbf \lambda_{c,r})$ on the first equation above by the values of $\lambda_{c,w}$ and $\lambda_{c,r}$ obtained from the second and third equations we get
\begin{equation}\label{eq:mddotxfext}
\mathbf{M} \,
\ddot{\mathbf{x}} -\mathbf{J}_{c,w}^T \, T + \mathbf{J}_{c,r}^T \mathbf B_r + \mathbf{J}_{ext,b}^T F_{ext} = \mathbf{0}.
\end{equation}
In the absence of external forces, we can solve for the torques
\begin{equation}
\mathbf T \big\rvert_{\mathbf F_{ext}=0} = \mathbf{J}_{c,w}^{-T} \Big[\mathbf M \ddot{\mathbf{x}} + \mathbf{J}_{c,r}^T \mathbf B_r \Big].
\end{equation}
\begin{figure}\centering
\includegraphics[width=82mm, clip=true ] {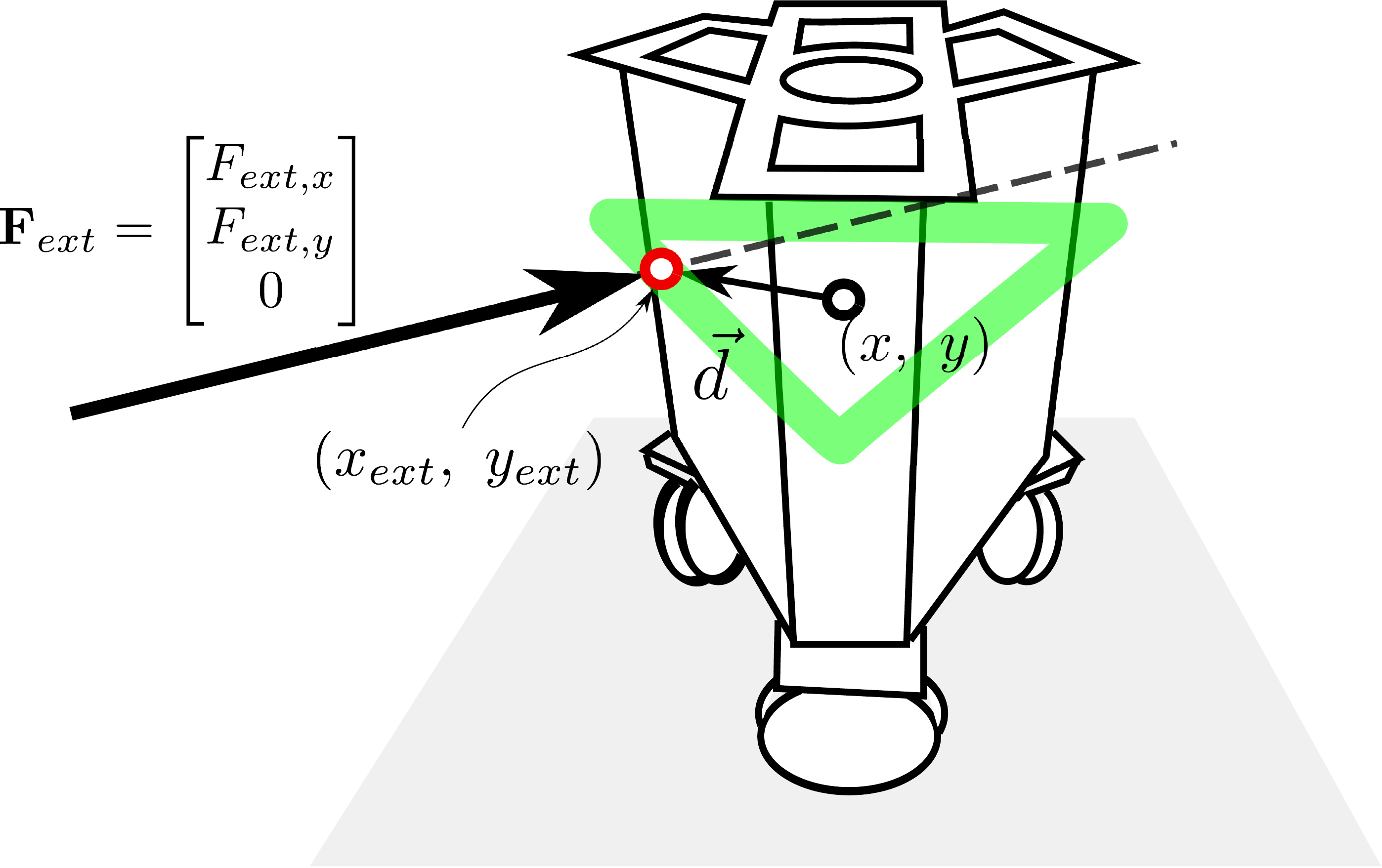}
\caption{{\bf External Force Estimation} is predicated on the assumption that the external force is a purely translational push applied to the robot's surface, as approximated by a triangular prism. The green triangle is the approximated robot body shape in a horizontal plane, and the perceived contact point, a red circle, occurs at the first of two intersections between this triangle and the line of zero external moment.}
\label{ext_force}
\end{figure}
The important point of the mapping above is that it can be numerically solved using the model and the acceleration estimate of Eq. (\ref{eqddotx}). On the other hand, when the robot collides with the environment, the torque sensors read values according to the dynamics of Eq. (\ref{eq:mddotxfext}). Assuming the output torque is equal to the value given by the torque sensors, i.e. $\mathbf T_s = \mathbf T$, we can use the previous two equations to solve for the external forces
\begin{equation}
\left(\mathbf T \big\rvert_{\mathbf F_{ext}=0} - \mathbf T_s\right) = \mathbf{J}_{c,w}^{-T} \mathbf{J}_{ext,b}^T F_{ext},
\end{equation}
which can be written in the alternative form
\begin{equation}\label{eq:alter}
\mathbf{J}_{ext,b}^T F_{ext} = \mathbf{J}_{c,w}^T \left(\mathbf T \big\rvert_{\mathbf F_{ext}=0} - \mathbf T_s\right).
\end{equation}
We now make the following simplifying assumptions:
\begin{itemize}
\item The external wrench has no net torque.
\item The external wrench is applied at a point on the triangular prism approximation of the body
\item The external wrench is always of a pushing nature
\end{itemize}
With those premises and the expression of Eq. (\ref{eq:jext}), the above equation becomes
\begin{multline}\label{f_ext04}
\Big[
F_{ext,x}, \;\;
F_{ext,y}, \;\;
\left(x_{ext} - x \right) F_{ext,y} -\left(y_{ext} - y \right) F_{ext,x}
\Big]^T\\
= \mathbf{J}_{c,w}^T \left(\mathbf T \big\rvert_{\mathbf F_{ext}=0} - \mathbf T_s\right).
\end{multline}
This equation has four unknowns, $\{F_{ext,x}, F_{ext,y}, x_{ext},$ $y_{ext} \}$ but only three entries. It is attempting to simultaneously solve the external force and its location. Let us focus on the third entry of the above equation. The third row can be written in the form
\begin{equation}\label{eq:line}
\left(x_{ext} - x \right) F_{ext,y}-\left(y_{ext} - y \right) F_{ext,x}
=
I_b\, \ddot{\theta} - \frac{R}{r_w} \sum_{i=0}^{2} \tau_{s,i}.
\end{equation}
This derivation comes from first comparing Eqs. (\ref{eq:mddotxfext}) and (\ref{eq:alter}), which lead to
\begin{equation}
\mathbf{J}_{c,w}^T \left(\mathbf T \big\rvert_{\mathbf F_{ext}=0} - \mathbf T_s\right) = \mathbf{M} \,
\ddot{\mathbf{x}} -\mathbf{J}_{c,w}^T \, T + \mathbf{J}_{c,r}^T \mathbf B_r,
\end{equation}
and then deriving the third row of the right hand side of the above equation, yielding
\begin{equation}
\mathbf{J}_{c,w}^T \left(\mathbf T \big\rvert_{\mathbf F_{ext}=0} - \mathbf T_s\right)\bigg\rvert_{row \; 3}
= I_b \ddot \theta - \frac{R}{r_w} \sum_{i=0}^{2} \tau_{s,i}.
\end{equation}
The above results are obtained from the third rows of the transpose of Eqs. (\ref{eq:jcw}) and (\ref{eq:jcr}), i.e.
\begin{gather}
J_{c,w}^T\big\rvert_{row \; 3} = \frac{1}{r_w} 
\begin{pmatrix}
R&&R&&R
\end{pmatrix},\\
J_{c,r}^T\big\rvert_{row \; 3} =
\begin{pmatrix}
0&&0&&0
\end{pmatrix}.
\end{gather}
Because Eq. (\ref{eq:line}) corresponds to a geometric line, the location of the contact point can be solved using solely Eq. (\ref{eq:line}) and our previously stated assumptions. The line is parallel to the direction of the external force, $\mathbf{F}_{ext}$, and can be used to find the distance from the center of the robot to the intersection of the line with the robot's body. The shape of our mobile base can be approximated as a triangular prism, and its planar section is a triangle, which is convex. Thus, there are only two points on its body where the line meets the premises. Therefore, we solve for the location where the external force is applied using those geometric constraints as shown in Fig. \ref{ext_force}.

Once we find the location of the contact point, we now solve for the external force using the first and second row of Eq. (\ref{f_ext04}).

\begin{figure*}[ht]
\centering
\includegraphics[width=\textwidth]{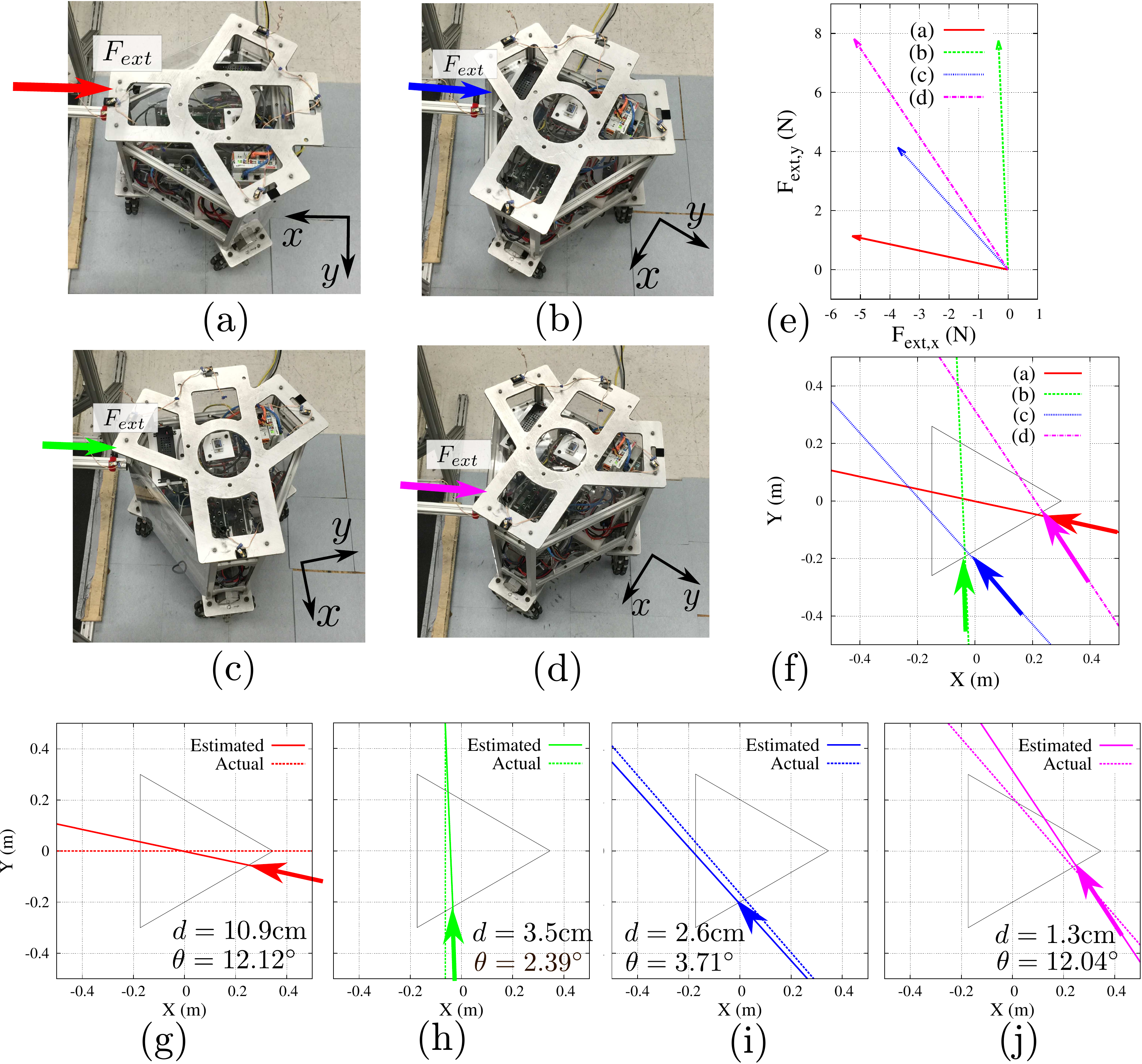}
\caption{{\bf Estimating Static External Forces} using only the torque sensors results in an accurate estimate of their location, angle, and magnitude. In Subfigures (a-d), the bar on the left side of the image confers an external push of roughly 9 Newtons onto the robot, above which is overlaid a triangle and a dot. This overlay is meant to reveal the robot's internal coordinate system, for clarity. The estimated forces from all four robot positions are shown directionally in Subfigure (f), where they are represented in the coordinate system of the robot. Subfigure (e) illustrates the magnitude of these forces, in the same coordinate system, emanating from the origin.}
\label{detection}
\end{figure*}

\section{Experimental Results and Assessment}


Throughout the previous sections we have established the following infrastructure: (1) full-body collision detection capabilities using constrained models and including wheel and side roller dynamics; (2) estimation of roller damping which is dominant in the behavior of the output robot dynamics; (3) fast collision response capabilities by achieving desired impedances through an admittance controller; (4) an experimental infrastructure including, a mobile base with torque sensors on the wheel drivetrains, a calibrated collision dummy, and a motion capture system.

The goal of this section is multi-objective: (1) to characterize the performance of our infrastructure in terms of accuracy of force detection and the impact location, (2) to measure the amount of time that takes our robot to detect collisions, (3) to measure the amount of time it takes our robot to respond to collisions once they have been detected, (4) to poke the robot in various places to proof that we can detect collisions in all parts of the robot including its wheels, and (5) to give an idea of what are the implications of our methodology for providing safety in human-scale mobile bases.

To do so, we conduct five calibrated experiments where we measure performance using a combination of the wheel torque sensor data, the wheel odometry and the motion capture data on the robot and the collision dummy. Additionally, we conduct a proof of concept experiment on safety, where the robot roams freely around people in all sorts of postures and collides with them safely.

\subsection{Detection of External Force and Contact Location}
In this experiment we evaluate our method's ability to detect the point of contact on the robot's body, the direction of the external force, and the magnitude of the external force. In particular we will use only the wheel drive-train torque sensors to identify those quantities without any use of external sensor mechanisms. In other words, the robot does not utilize motion capture data or wheel odometry to detect those quantities. 

\begin{figure*}[ht]\centering
\includegraphics[width=\linewidth, clip=true]{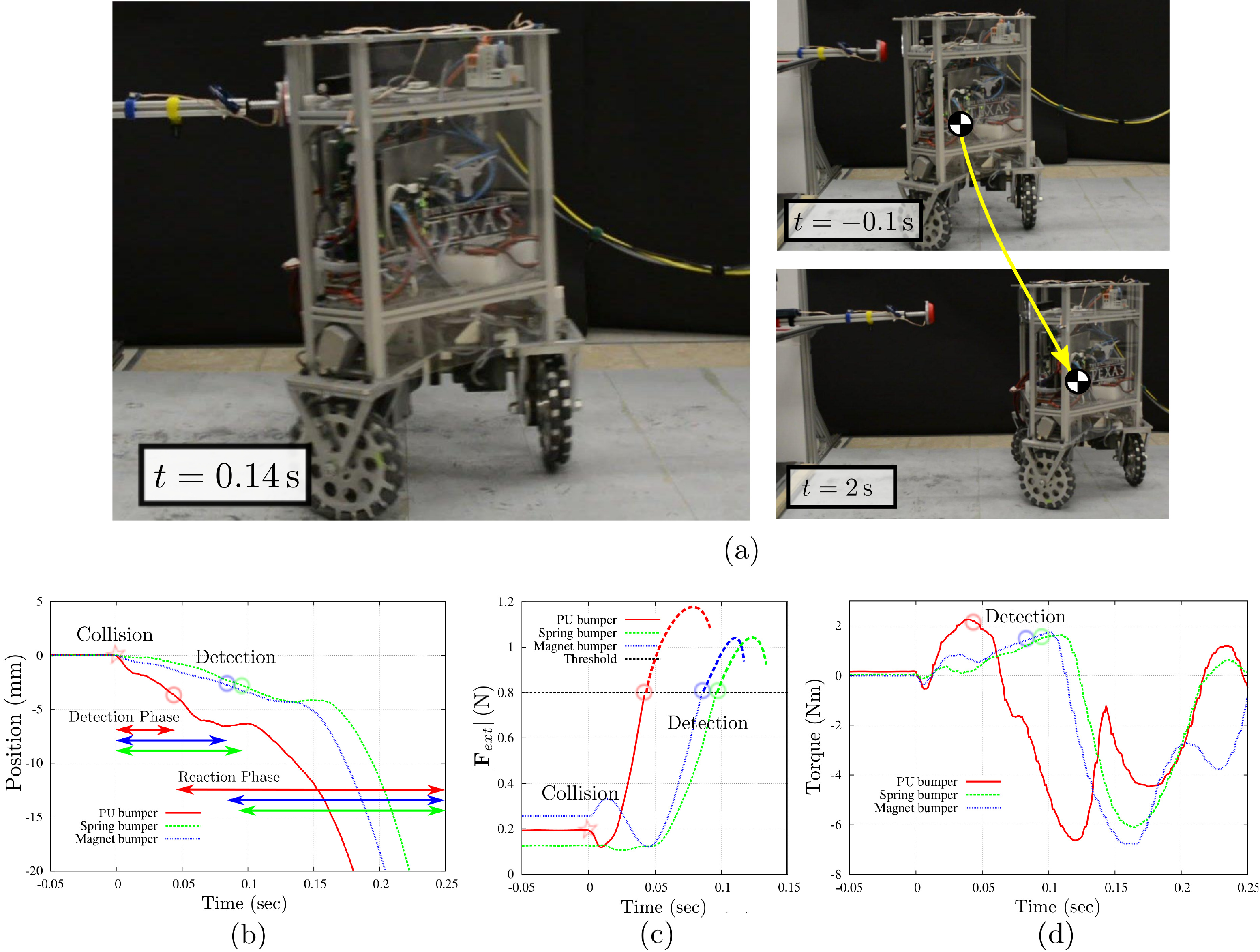}
\caption{{\bf Upper Body Collision Testing} illustrates the robot's collision avoidance behavior with respect to the three different bumper designs when the impact occurs above its center of mass. The magnet bumper impact is shown at three representative frames in (a), with $t=0.14 \rm s$ representing the peak of force and spring deflection. Subfigure (b) shows the position evolution of the robot from the instant of contact, and highlights the instant when the robot's software registered the impact in each of the three trials. After the initial impact but before the robot recognized the impact, that is, during the detection phase, the force of impact pushed Trikey backwards. In (c), the measured torque sensor value on the Wheel 0 for each trial is plotted against the same time range. Note that the initial dip in torque is due the propensity of an upper body impact to tip the robot over, rolling the wheels forward. By virtue of being a more direct transfer of energy, the PU bumper is detected first, causes more initial motion, and results in a higher peak torque than the other experiments The faster detection is due to the larger torque, since external force measurement is based on a moving average filter of the torque sensor signals.}
\label{high_static_collision}
\end{figure*}

To conduct these tests, we use the infrastructure depicted in Fig. \ref{setup}. The horizontally sliding dummy is connected to a pulley system that runs to an overhead system with a vertical weight of $1$Kg. As a result a constant force of $10$N is applied to the slider. In Fig. \ref{detection} we show images of the experimental setup where the slider is placed in contact with the base before conducting the estimation process. The robot's wheels are  powered off, and because of the high friction of the harmonic drives, the forces applied by the dummy are not enough to push the robot away. 

Subfigs. \ref{detection} (a)-(d) show the procedure that we conduct. We first place the robot in different directions and orientations with respect to the dummy. Using only the torque sensor data, we proceed to use the force estimation techniques described in Sec. \ref{sec:forceestimation} to identify the point of contact, the direction of the force, and its magnitude. We repeat the same experiment for 4 different scenarios applying the same amount of force. Without loss of generality, all the external forces are applied to the same side of the robot as the robot is symmetrical.

Fig. \ref{detection} (e) and (f) shows the results of the estimation process. Subfig. \ref{detection} (e) shows that the magnitude and direction of the estimated forces and Subfig. \ref{detection} (f) shows the contact point and the force direction with respect to the base geometry and orientation. The magnitude of the forces estimated ranges from $5.5$N to $10$N. Those values are $(0\%-45\%)$ smaller than the $10$N of force applied by the contact dummy. We believe that the reason is due to stiction of the overall mechanical structures standing between the contact point, the wheels in contact with the ground, and the pulley system connecting the wheel to the torque sensors. The maximum error in detecting the direction of the forces is $3.3\%$ with respect to the full circle, or equivalently $12$deg over $360$deg with a mean value of $\pm 2\%$. Finally, the maximum error in detecting the point of contact is $11$cm with a mean value of $4.5$cm, or equivalently, $18\%$ of error with a mean value of $7.5\%$ with respect to the $61$cm of length of the robot's side walls. 

Overall we accomplish maximum errors of $45\%$ for the magnitude, $3.3\%$ for the direction and $18\%$ for the location of the external forces. The good accuracy of the location and direction of the estimated force can be leveraged to respond safely to impacts by moving away from the colliding bodies with precision. The medium accuracy of the estimated force's magnitude is probably due to the mechanical structure and not due to the estimation strategy. Nonetheless, it is sufficient for the controller to execute the admittance control model. However, if we wish to achieve the target impedance with high precision, the external force's magnitude will have to be estimated with higher accuracy. In that case improved designs of the mobile robot that minimize stiction should be sought.

\subsection{Collisions with Motionless Robot}
In this experiment we evaluate our method's ability to not only detect collisions but to quickly react in a manner that is perceived as safe. Moreover, the tests discussed here will analyze collisions with the mobile base standing motionless close to the collision dummy. Responding safely to collisions when the robot is still is one of the hardest case scenarios that a robot may encounter. In such case, the safe response of the robot solely depends on its ability to estimate the external forces with accuracy. In contrast, when a robot collides while in motion its controller knows the trajectory where it came from. As such a simple safe response would be to reverse direction towards that trajectory. 

Once more we use the infrastructure of Fig. \ref{setup}. However, this time around we connect the pulley system to a vertical weight of $4.54$Kg producing a constant horizontal force of $44.54$N on the contact dummy. The contact dummy is also now connected to a sliding weight of $9.08$Kg which constitutes the effective mass that collides with the mobile base. The sliding dummy is released at a certain distance to the robot and when it collides with the robot it has reached a velocity of $0.5m/s$. The robot is initially at rest and when it detects contact it moves away from the collision in the direction of the collision.

\begin{figure*}[ht]\centering
\includegraphics[width=\linewidth, clip=true ] {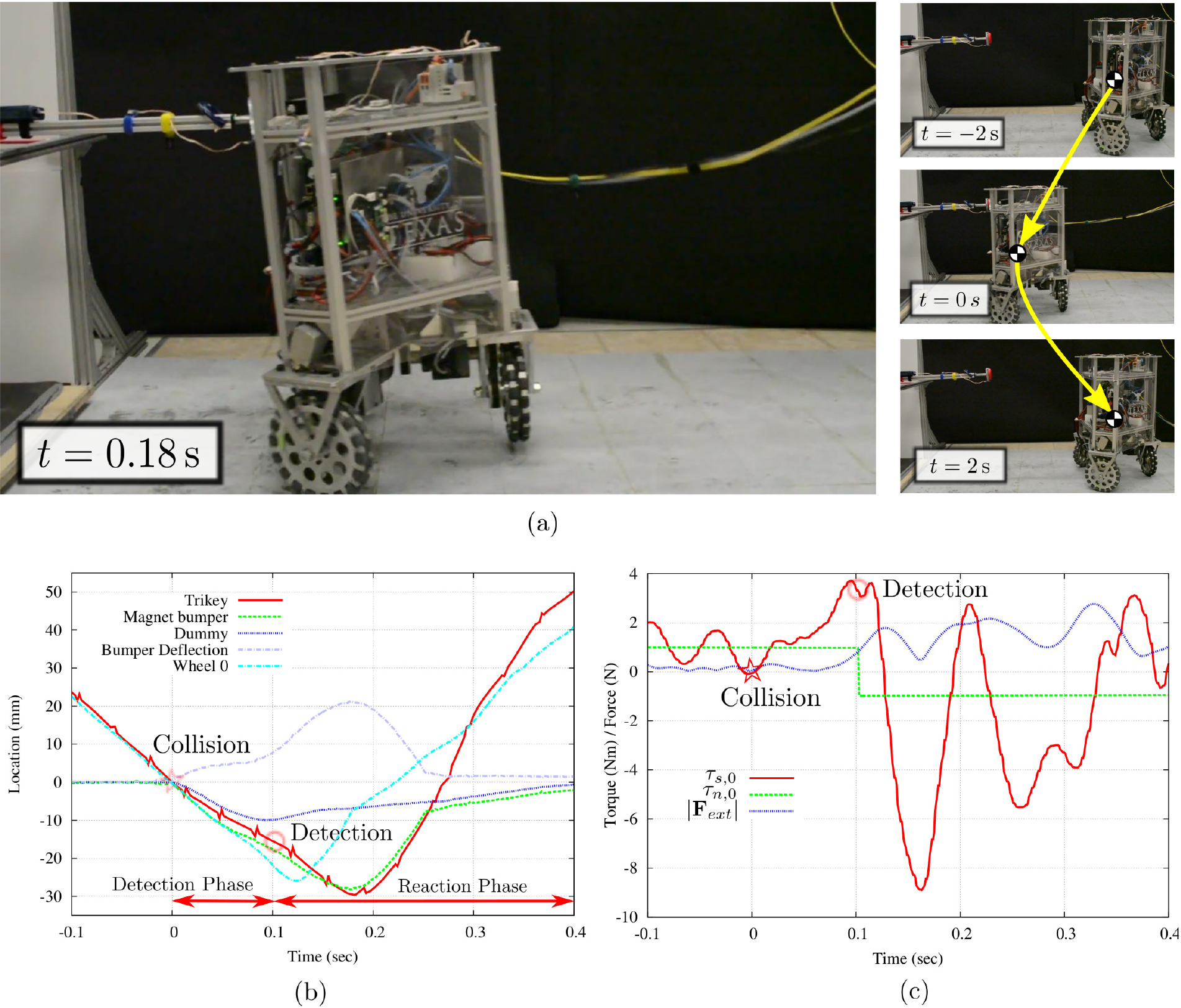} 
\caption{{\bf Collision Against a Static Obstacle} tests Trikey's ability to reverse direction when moving at full speed, after an impact with the magnet bumper. Various stills in (a), including the $t=0.18 \rm s$ frame with maximum spring deflection, illustrate the experimental procedure. Subfigure (b) plots various reference positions including the position of Trikey itself, the position of the bumper, the position of the slider, the spring deflection, and the angle of Wheel 0 (times a scaling factor), with all positions normalized to zero at the instant of collision.
Subfigure (c) plots the torque sensor from Wheel 0, the expected Wheel 0 torque sensor value, and the estimated external force. This external force exceeds the predefined collision threshold when $t= 105\rm ms$, corresponding to the Detection timestamps in both (b) and (c).}
\label{high_moving}
\end{figure*}

In Fig. \ref{high_static_collision} we show the procedure that we conduct. We first place the robot next to the collision dummy with the dummy separated from the robot. Once more, we only use torque sensor data to estimate the direction, location and magnitude of the collision and respond to it. We implement the force estimation procedure of Sec. \ref{sec:forceestimation} and the admittance controller of Sec. \ref{sec:syschar}. The desired impedance that we implement for the controller is $M_{des} = 2kg$ and $B_{des} = 1.6N/m^2$. The motivation for these values is first to maximize the reaction speed by setting a low target mass. However, if we make $M_{des}$ too small, the robot accelerates too quickly in reaction to the collision and it tips over. Therefore, we decrease it to just over the limit where it tips over. In order to select $B_{des}$ we follow the subsequent procedure. Using Eq. (\ref{eq:traj}), the position achieved by the controller on a particular direction, e.g. $x$, after impact at time $\infty$ is
\begin{equation}\label{eq:xdes}
x_{des}(t\rightarrow \infty) = x_0 + \frac{F_{ext,x}}{B_{des}}.
\end{equation}
Based on this equation, we design $B_{des}$ such that the robot moves away by $0.5$m upon collision, i.e. 
\begin{equation}
x_{des}(t\rightarrow \infty) = x_0 + 0.5.
\end{equation}
Taken into account that we use a threshold of $|\mathbf F_{ext}| = 0.8N$ to initiate the admittance controller (see Fig. \ref{controller}), solving Eq. (\ref{eq:xdes}) for these values we get $B_{des} = 0.8/0.5 = 1.6 N/m^2$.

We conducted the collision experiments using three different materials on the collision dummy: the default thin polyurethane plastic (PU bumper), a thin polyurethane foam with a spring (Spring bumper), and the same thin polyurethane foam with the spring and a magnetic latch as described in Subsection \ref{material} (Magnetic bumper).

\begin{figure*}[ht]
\centering
\includegraphics[width=0.95\textwidth] {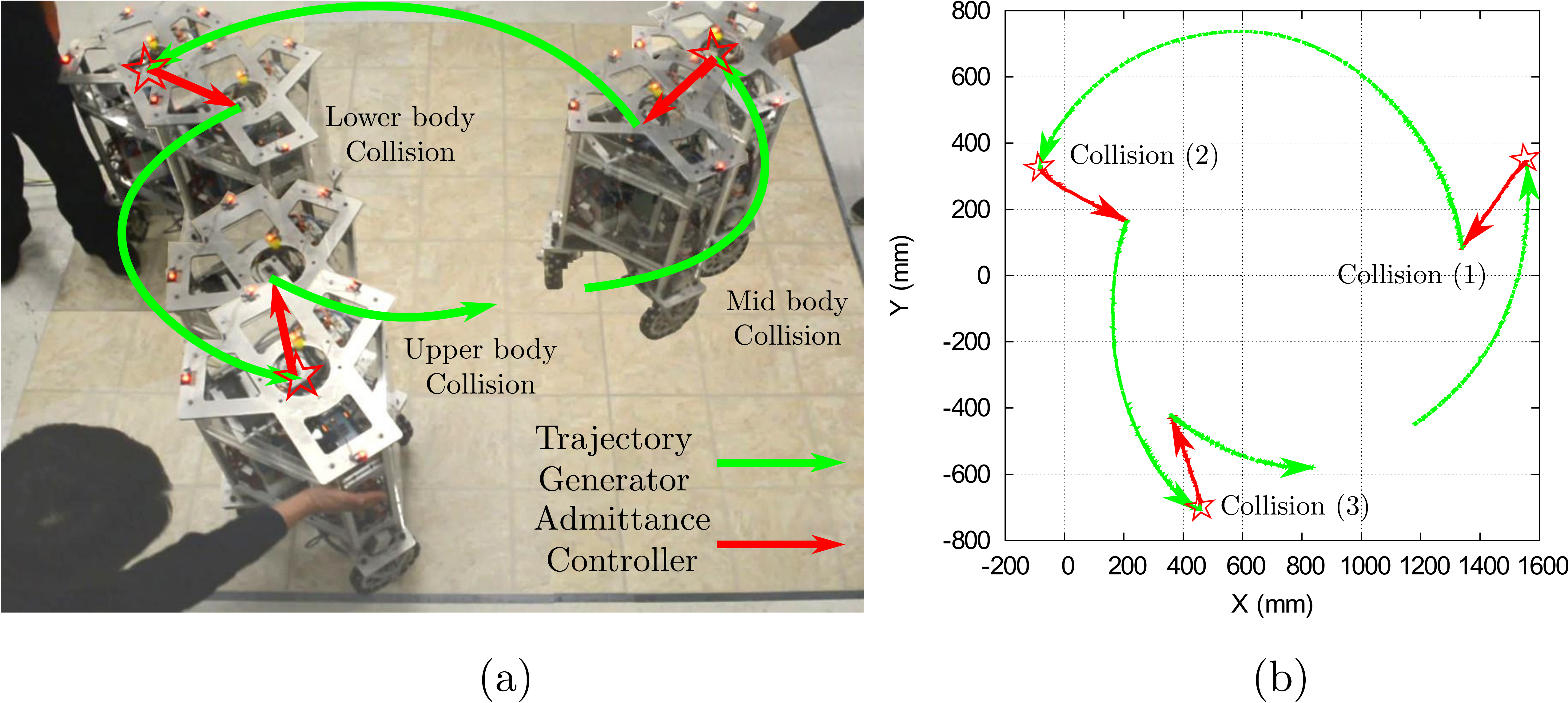} 
\caption{{\bf Omni-directional Motion with Unplanned Collisions} demonstrates Trikey's full motion capability as it moves about a $1.5 \rm m$ diameter circle at $0.16 \rm m/s$. A composite image of several frames, (a) shows the motion, the escapes, and the human obstacles in the experiment. The trajectory captured by the motion capture system is shown in (b).}
\label{omni_collision}
\end{figure*}

Additionally, to compare performance, we conduct collision tests both in the upper and lower parts of the mobile base. As shown in Fig. \ref{high_static_collision} (c), the collision was detected in $45$ms (PU bumper), $95$ms (Spring bumper), and $85$ms (Magnet bumper). 

In Subfig. \ref{high_static_collision} (c) we observe that that the estimated external force in the PU bumper case reaches the reaction threshold at $t=0.05$s. As a result, the admittance controller kicks in causing the robot to move quickly away. As shown in Subfig. \ref{high_static_collision} (b), after detecting the contact, the robot's change in position seems to hit a plateau for about $50$ms. The reason is due to the robot accelerating quickly and lifting the front wheel (see Subfig. \ref{high_static_collision} (a) for that effect). After that plateau, the robot quickly moves away from the collision.

Let us focus on Subfig. \ref{high_static_collision} (d). Positive wheel torques result from the impact forces on the robot and negative torques result from the robot moving away from the impact. As we can see, using the spring and magnet bumpers reduces the impact torques by about $20\%$ with respect to the peak value of the PU bumper. Additionally, if we focus on Subfig. \ref{high_static_collision} (b) we can see during the collision time, $t\in(0,t_{detection})$, the robot's trajectory associated with the response to the PU bumper accelerates much more quickly than that for the spring or magnetic bumpers. It is this combination of lower peak force and lower acceleration that will make the use of the spring or magnetic bumpers safer.



\subsection{Collisions with Moving Robot}
The setup for this experiment is similar to the one before. However, the robot now moves towards the resting contact dummy and produces a collision to which it needs to respond. This experiment tests the reaction time and peak torques of the moving robot upon collision.

The collision dummy is initially at rest with a total sliding mass of $13.62$Kg. The robot moves towards the dummy and hits it with a velocity of $0.22 \rm m/s$. This time around, we only conduct the experiment with the magnetic bumper. The same estimation and control methods used in the previous section are applied.

Similarly to the tests before that contain the spring or magnetic bumper, it takes $105$ms to detect the collision threshold. Fig. \ref{high_moving} (c) shows the measured torque from the torque sensor in Wheel 0 and the magnitude of the estimated external force. 

Overall, the reaction time is similar to the motionless experiment before and the peak torque values are about twice the values we had obtained with the spring or magnetic bumpers. This increase in value is due to the robot having an initial velocity which causes a higher force collision due to the robot's heavy weight.

\subsection{Additional Experiments}

In Fig. \ref{omni_collision} an experiment involving our mobile base executing circular arc trajectories while being pushed away is presented. The trajectory of the mobile base is recorded using the motion capture system. As we can see, collisions are promptly detected resulting in the robot reacting to them in the opposite direction of the colliding force.

\begin{figure*}[ht]\centering
\includegraphics[width=0.85\textwidth] {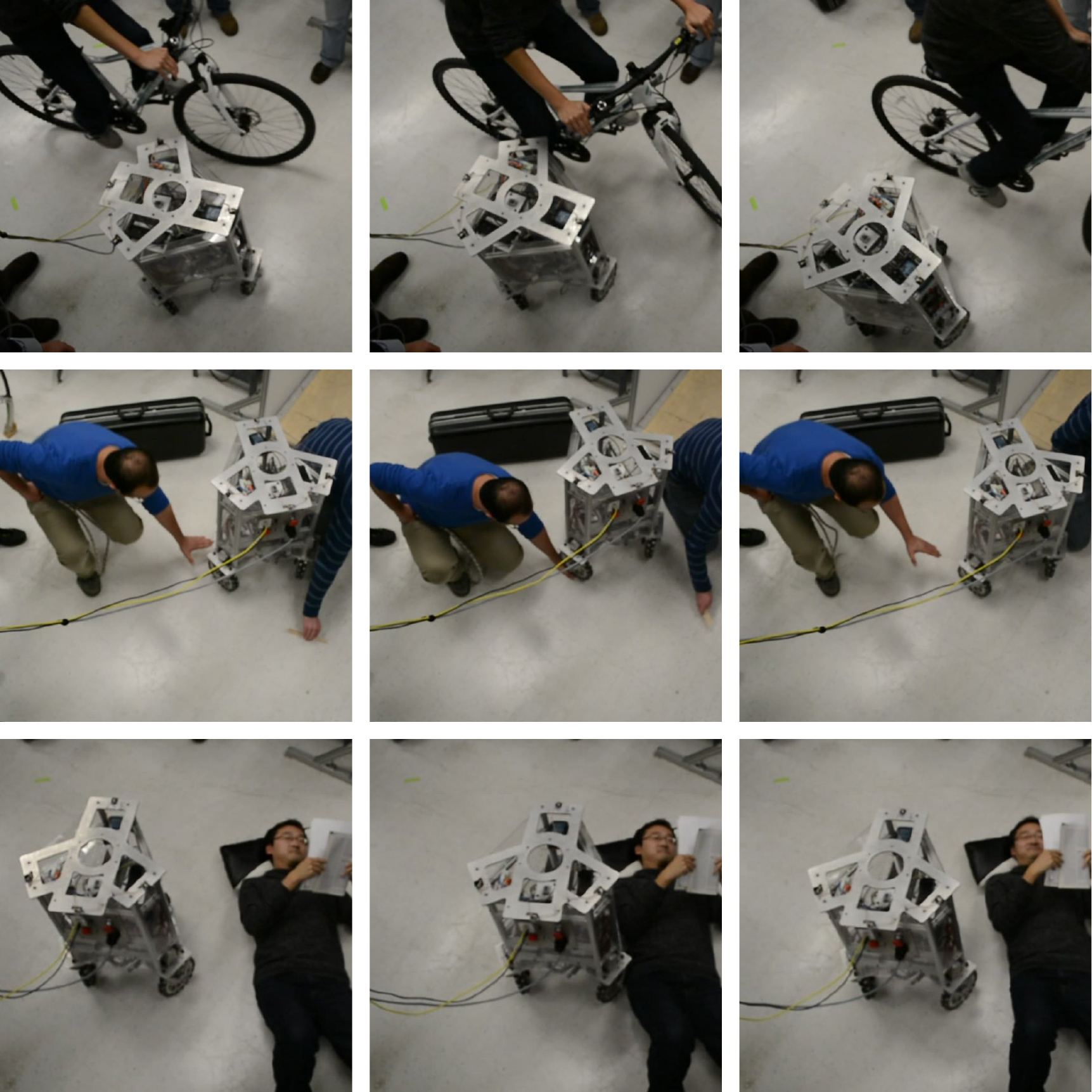}
\caption{{\bf Collisions in Human Environments} points towards our long term vision for mobile robots. Here Trikey collides with humans in various scenarios, and reacts to the collisions safely.}
\label{collisions}
\end{figure*}

Finally, as a proof of concept, we conducted an experiment where we let the mobile base move around performing circular arc trajectories while people provide it with simulated accidental collisions. Fig. \ref{collisions} shows the robot's reaction to collision with a bicycle, a hand placed on the floor in the wheel path, and a person lying down.

\section{Conclusions}

Mobile robots will not be truly useful until they are very safe in cluttered environments. We have presented a methodology for these types of robots to quickly react and achieve low impedance behaviors upon discovering an unexpected collision. It is the first study to accomplish full-body collision detection on all parts of a mobile platform.

Our estimation method has been shown to estimate the contact location of the collisions with $18\%$ error, direction of the contact forces with $3.3\%$ error, and magnitude of those forces with $45\%$ error. The lower accuracy of the magnitude is due to mechanical limitations of the structure of the base and the connection of the wheel to the torque sensors. Those could be improved by having a stiffer structure and improved connections from the wheel to the torque sensor.

Empirically estimating roller dynamics has been key to enhancing the external force sensing accuracy. We have chosen to use only model based estimation and have achieved good precision but feel we could benefit in the future from statistical methods such as \cite{Fang2011}. Our detection and reaction to collisions rely solely on the on-board torque sensor data. They do not rely on wheel odometry or external global pose estimation. As such they can attain a very fast reaction time.

As shown in the experiment of Fig. \ref{omni_collision}, the admittance controller developed in Sec. \ref{sec:syschar} has been effective at providing the desired impedances. In particular, it decreased the peak contact torques without tipping over the base. At the same time, the desired closed loop damping prevents the robot from moveing too far away from the collision source. These parameters can be tuned depending on the application.


In the future we would like to conduct experiments with test dummies that are clamped against a wall. Such scenario is one of the most dangerous ones. We would also like to apply safety criteria that compare the static and dynamic forces of our base to the maximum tolerable curves obtained from previous empirical studies. Additionally, we would like to study collisions of the base at moderately high speeds. Because bases are heavy and have limited braking ability, their reaction capabilities are similar to those of cars. As such we would like to apply injury indicators from the automotive industry to explore these types of collisions.

\bibliographystyle{plain}      
\bibliography{bib}   

\end{document}